\documentclass{article}

\usepackage[nonatbib,preprint]{neurips_2021}

\usepackage[utf8]{inputenc} 
\usepackage[T1]{fontenc}    
\usepackage{hyperref}       
\usepackage{url}            
\usepackage{booktabs}       
\usepackage{amsfonts}       
\usepackage{nicefrac}       
\usepackage{microtype}      
\usepackage{xcolor}         

\usepackage{graphicx}
\usepackage[maxbibnames=99]{biblatex}
\usepackage{textcomp,xspace}
\usepackage{dsfont}
\usepackage{amsmath}
\usepackage{svg}
\usepackage{subcaption}
\usepackage{multirow}
\usepackage{enumitem}
 \usepackage{diagbox}

\newcommand{\taskset}{$\mathcal{T}$\xspace}
\newcommand{\src}{$T_{src}$\xspace}
\newcommand{\tgt}{$T_{tgt}$\xspace}

\newcommand\Tstrut{\rule{0pt}{2.6ex}}         
\newcommand\Bstrut{\rule[-0.9ex]{0pt}{0pt}}   

\title{Zero-shot Task Adaptation using Natural Language}

\author{%
  Prasoon Goyal, Raymond J. Mooney, Scott Niekum \\
  Department of Computer Science\\
  University of Texas at Austin\\
  \texttt{\{pgoyal,mooney,sniekum\}@cs.utexas.edu} \\
    
}

\bibliography{neurips}

\begin{document}

\maketitle

\begin{abstract}
Imitation learning and instruction-following are two common approaches to communicate a user's intent to a learning agent. However, as the complexity of tasks grows, it may be beneficial to use both demonstrations and language to communicate with an agent.
In this work, we propose a novel setting where, 
given a demonstration for a task (the \emph{source task}), and a natural language description of the differences between the demonstrated task and a related but different task (the \emph{target task}), our goal is to train an agent to complete the target task in a zero-shot setting---that is, without \emph{any} demonstrations for the target task.
To this end, we introduce Language-Aided Reward and Value Adaptation (LARVA) which, given a source demonstration and a linguistic description of how the target task differs, learns to output either a reward or value function that accurately reflects the target task.
Our experiments show that on a diverse set of adaptations, our approach is able to complete more than 95\% of target tasks when using template-based descriptions, and more than 70\% when using free-form natural language.
\end{abstract}

\section{Introduction}

Teaching learning agents how to perform a new task is a central problem in artificial intelligence. One paradigm, namely imitation learning \cite{argall2009survey}, involves showing demonstration(s) of the desired task to the agent, which can then used by the agent to infer the demonstrator's intent, and hence, learn a policy for the task. However, for each new task, the agent must be given a new set of demonstrations, which is not scalable as the number of tasks grow, particularly because providing demonstrations is often a cumbersome process. 
On the other hand, techniques in instruction-following \cite{macmahon2006walk,vogel2010learning,chen2011learning} communicate the target task to a learning agent using natural language. As the complexity of tasks grow, providing intricate details using natural language could become challenging.

\begin{figure}
    \centering
    \includegraphics[width=0.8\textwidth]{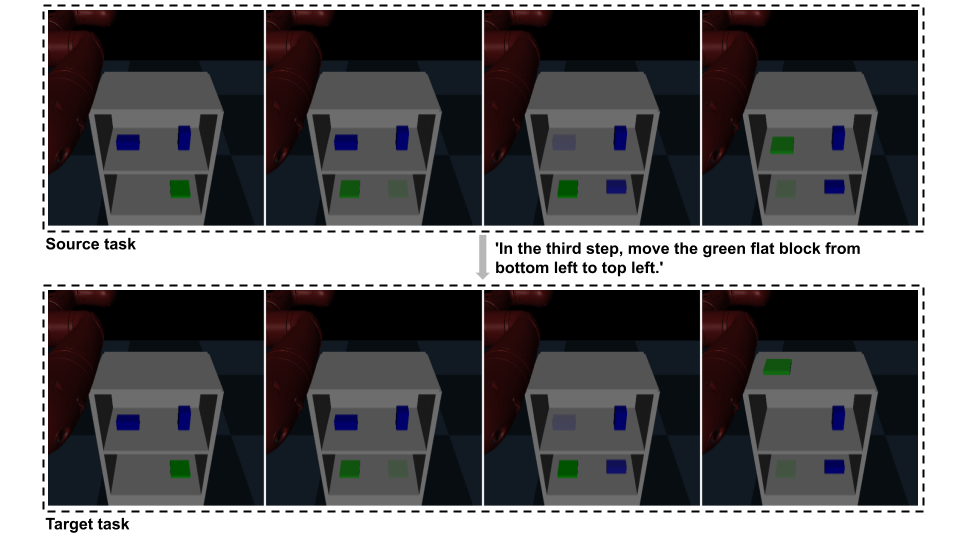}
    \caption{Example of the setting: The top row shows the \emph{source task}, while the bottom shows the \emph{target task}. Given a demonstration of the source task, and a natural language description of the difference between the two tasks such as ``In the third step, move the green flat block from bottom left to top left.'', our goal is to train an agent to perform the target task \emph{without} any demonstrations of the target task.}
    \label{fig:environment}
\end{figure}

This motivates a new paradigm to teach agents, that allows scaling up learning from demonstration to multiple related tasks with a single (or a few) demonstration(s) by using a much more natural modality, namely, language. At the same time, intricate details which are harder to communicate using language alone can be provided using demonstration(s).
To this end, we propose a novel setting---given a demonstration of a task (the \emph{source task}), we want an agent to complete a somewhat different task (the \emph{target task}) in a \textbf{zero-shot} setting, that is, without access to \emph{any} demonstrations for the target task. The difference between the source task and the target task is communicated using language. 
The proposed setting requires combining information from both the demonstration and the language, and can therefore serve as an important step towards building systems for more complex tasks which are difficult to communicate using demonstrations or language alone.

For example, consider an environment consisting of objects on different shelves of an organizer, as shown in Figure~\ref{fig:environment}. Suppose the \textbf{source task} (top row) requires moving the green flat block from bottom-right to bottom-left, the blue flat block from middle-left to bottom-right, and then the green flat block from bottom-left to middle-left. The \textbf{target task} (bottom row) is similar, but in the final step, the green flat block should be moved to top-left instead. We posit that given a demonstration for the source task, and a free-form natural language description of the difference between the source and the target tasks, 
such as ``In the third step, move the green flat block from bottom left to top left'', 
an agent should be able to infer the goal for the target task. We propose a framework that can handle a diverse set of adaptations between the source and the target tasks, such as a missing step, an extra step, and swapping the final positions of two objects.

The environment has a similar structure to several real-world applications, where task adaptation using language could be particularly beneficial. For instance, consider the goal of building service robots for home environments. These robots must be able to learn a wide variety of tasks from non-expert users. Many tasks, such as cooking or assembly, involve a sequence of discrete steps, and such tasks could have several variations, like different cooking recipes or assembling different kinds of furniture. Being able to demonstrate one (or a few) of these tasks, and then communicating the difference between the demonstrated task(s) and other similar tasks could significantly reduce the burden of teaching new skills for the users.

These problems involve planning/control at 2 levels---high-level planning over the steps, and low-level control for executing each step. Since our proposed algorithm focuses on the high-level planning, we illustrate our approach on the simple environment shown in Figure~\ref{fig:environment}, where the low-level control is abstracted away. However, our framework is general, and can be combined with approaches that perform low-level control.


The proposed setting is challenging for several reasons. 
First, most existing approaches in imitation learning and instruction-following infer the goal for a target task from a demonstration or an instruction, respectively. However, in our setting, neither of these modalities is sufficient by itself, and the agent must be able to combine complementary information from the source demonstration(s) and the natural language descriptions, in order to infer the goal for the target task.
Second, in order to understand the natural language description, the agent must be able to map concepts in the description to objects and actions, a problem known as symbol grounding \cite{harnad1990symbol}.
Finally, in order to be scalable, we intend to learn a purely data-driven model that can does not require engineering features for the language or the environment, and can learn to infer the goal for the target task directly from data.

We introduce the Language-Aided Reward and Value Adaptation (LARVA) model that takes in a dataset of source demonstrations, linguistic descriptions, and either the reward or optimal value function for the target task, and is trained to predict the reward or optimal value function of the target task given a source demonstration and a linguistic description. 
The choice between reward and value prediction could be problem-dependent---for domains with complex transition dynamics, learning a value function requires reasoning about these dynamics, and therefore, it might be better to use LARVA for reward prediction, with a separate policy learning phase using the predicted rewards; for domains with simpler dynamics, a value function could be directly predicted using LARVA, thereby removing the need for a separate policy learning phase.

We experiment with a diverse set of adaptations, and show that the model successfully completes over 95\% target tasks when using synthetically generated language, and about 75\% target tasks when using unconstrained natural language collected using Amazon Mechanical Turk.


\section{Related Work}


\paragraph{Imitation Learning.}
Imitation learning is one of the standard paradigms for teaching new skills to learning agents. 
The agent is provided with demonstration(s) of a task, and must infer the demonstrator's intent, and hence, learn to complete the task \cite{argall2009survey}.
Approaches to imitation learning can broadly be classified into behavior cloning \cite{pomerleau1989alvinn,ross2010efficient,ross2011reduction}, inverse reinforcement learning \cite{abbeel2004apprenticeship,ramachandran2007bayesian,ziebart2008maximum,finn2016guided}, and adversarial learning \cite{ho2016generative,fu2017learning}.
Our proposed setting differs from standard imitation learning, since the agent is provided with demonstration(s) of the source task, but needs to infer the reward for a related but different target task, the difference being communicated using language.

\paragraph{Transfer Learning.}
Another closely related problem to our proposed setting is transfer learning, wherein an agent trained on a source task needs to complete a related but different target task.
The agent is \emph{finetuned} on data from the target task in transfer learning, and the objective is to reduce the amount of data needed for the target task by effectively transferring experience from the source task \cite{pan2009survey}. 
This is different from our proposed setting, because in our setting, we don't need to transfer \emph{experience} from the source task to the target task; instead, the demonstration(s) for the source task must be used with the description to infer the goal for the target task. 

\paragraph{Meta-learning and Few-shot Learning.}
Our setting is also related to the meta-learning and few-shot learning settings  \cite{vanschoren2018meta,wang2020generalizing}. 
In meta-learning, an agent is given training data from multiple tasks sampled from a distribution of tasks, and it must use these data to generalize to new tasks sampled from the distribution.
The data from these tasks could be used to extract useful features, build models as a pretraining step, or learn a training routine (e.g. how to search over the space of neural network architectures), which can be transferred to the new task to learn from fewer datapoints, learn a more robust model, or converge to a solution faster.
Few-shot learning is a subcategory of techniques within meta-learning, where the goal is to learn a new task from few training examples.
Our approach infers the target reward function for a new target task given a source demonstration and a description in a zero-shot setting, which is a special case of few-shot learning.

\paragraph{Language as Task Description.}
In a large class of problems, which can broadly be termed as \emph{instruction-following}, language is used to communicate the task to a learning agent.
In this setting, the agent is given a natural language command for a task, and is trained to take a sequence of actions that complete the task.
A well-studied problem in this setting is the Vision-and-language Navigation, where the tasks consist of navigating to a desired location in an environment, given a natural language command and an egocentric view from the agent's current position \cite{anderson2018vision,fried2018speaker,wang2019reinforced}.
Another subcategory of instruction-following involves instructing an embodied agent using natural language \cite{tellex2011understanding,hemachandra2015learning,arkin2017contextual,shridhar2020alfred,stepputtis2020language,misra2016tell,sung2018robobarista}.
Our proposed setting is different from instruction-following, in that the goal of the target task is not being communicated using language alone; instead, a demonstration for a related task (source task) is available, and language is used to communicate the difference between the demonstrated task and the target task. Thus, the information in the demonstration and the language complement each other.

\paragraph{Language to Aid Learning.}
Several approaches have been proposed in the past that use language to aid the learning process of an agent. In a reinforcement learning setting, this could take the form of a language-based reward, in addition to the extrinsic reward from the environment \cite{luketina2019survey,goyal2019using,goyal2020pixl2r,kaplan2017beating}, or using language to communicate information about the environment to the agent \cite{wang2021grounding,branavan2012learning,narasimhan2018grounding}. 
\textcite{andreas2017learning} use language in a meta-learning setting for few-shot transfer to new tasks. While related to our setting, in this work, language is provided for each task independently, and tasks are deemed similar if their linguistic descriptions are related. In our setting, however, language is used to explicitly describe the difference between two tasks.


\section{Problem Definition}


Consider a \textbf{goal-based task}, which can be defined as a task where the objective is to reach a designated goal state in as few steps as possible.
It can be expressed using the standard Markov Decision Process (MDP) formalism, as
$M = \langle S, A, P, R, \gamma, g\rangle$,
where
$S$ is the set of all states,
$A$ is the set of all actions,
$P : S \times A \times S \rightarrow [0, 1]$ is the transition function,
$R : S \rightarrow \mathds{R}$ is the reward function,
$\gamma \in [0, 1]$ is the discount factor, and
$g \in S$ is the unique goal state.

At timestep $t$, the agent observes a state $s_{t} \in S$, and takes an action $a \in A$, according to some policy $\pi : S \times A \rightarrow [0, 1]$.
The environment transitions to a new state $s_{t+1} \sim P(s_{t}, a_{t}, \cdot)$, and the agent receives a reward $R_{t} = R(s_{t+1})$.
The objective is to learn a policy $\pi^{*}$, such that the expected future return, $G_{t} = \sum_{t'=t}^{t_{max}} \gamma^{t'-t} R_{t}$, is maximized.
Further, $V^{*}_{R} : S \rightarrow \mathds{R}$ denotes the optimal value function under the reward function $R$, and can be used to act optimally.


The reward function for a goal-based task can be defined as $R(s) = \mathds{1}[s=g]$, where $\mathds{1}[\cdot]$ is the indicator function.
Thus, for $\gamma < 1$, an optimal policy for a goal-based task must reach the goal state $g$ from any state $s \in S$ in as few steps as possible.



Let \taskset$=\{M_{i}\}_{i=1}^{N}$ be a \textbf{family} of goal-based tasks $M_{i}$, each with a distinct goal $g_{i}$, and the reward function $R_{i}$ defined as above.
The set of states $S_{i}$ and actions $A_{i}$, the transition functions $P_{i}$, and the discount factors $\gamma_{i}$ across different tasks may be related or unrelated \cite{kroemer2019review}.

For instance, in the environment shown in Figure~\ref{fig:environment}, a goal-based task consists of arranging the objects in a specific configuration defined by a goal state $g$,
while \taskset is the set of all multi-step rearrangement tasks in the environment.

Let \src, \tgt $\in$ \taskset be two tasks, and $L$ be a natural language description of the difference between the tasks.
Given a demonstration for the source task \src, and the natural language description $L$, our objective is to train an agent to complete the target task \tgt in a \textbf{zero-shot setting}, i.e., without access to the reward function or demonstrations for the target task.

\section{LARVA: Language-Aided Reward and Value Adaptation}
\label{sec:model}

We propose Language-Aided Reward and Value Adaptation (LARVA), a neural network that takes in a source demonstration, $\tau_{src}\xspace$, the difference between the source and target tasks described using natural language, $L$, and a state from the target task, $s \in S_{tgt}$, and is trained to predict either $R_{tgt}(s)$, the reward for the state $s$ in the target task, or $V^{*}_{R_{tgt}}(s)$, the optimal value of the state $s$ under the target reward function $R_{tgt}$.



We assume access to a training set $\mathcal{D} = \{(\tau^{i}_{src}, L^{i}, g^{i}_{tgt})\}_{i=1}^{N}$, where
$\tau^{i}_{src}$ is a demonstration for the source task of the $i^{th}$ datapoint,
$L^{i}$ is the natural language description of the difference between the source task and the target task for the $i^{th}$ datapoint, and
$g^{i}_{tgt}$ is the goal state for the target task.
The details of the dataset and the data collection process are described in Section~\ref{sec:data}.


Next, we describe the network architecture, and training details of LARVA.

\begin{figure}
    \centering
    \begin{subfigure}{0.8\textwidth}
        \includegraphics[width=\textwidth]{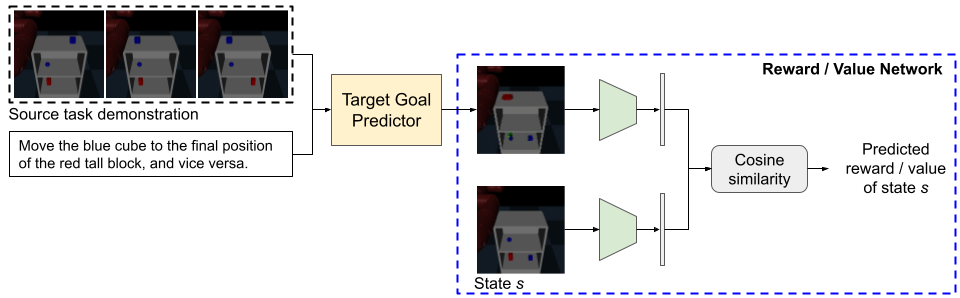}
        \caption{Full model: The target goal predictor takes in a source demonstration and a description to predict the goal state for the target task. The reward / value network takes this predicted goal state, and another state $s$ from the target task to predict the reward or value of the state $s$ under the target reward function.}
    \end{subfigure}
    \begin{subfigure}{0.8\textwidth}
        \includegraphics[width=\textwidth]{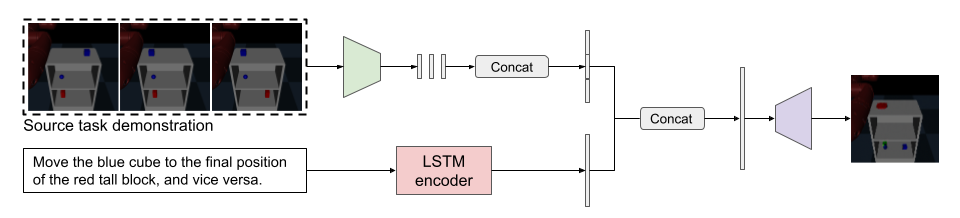}
        \caption{Target goal predictor}
    \end{subfigure}
    \caption{Neural network architecture for LARVA}
    \label{fig:nn-arch}
    \vspace{-10pt}
\end{figure}

\subsection{Network Architecture}


We decompose the learning problem into two subproblems: (1) predicting the goal state for the target task given the source demonstration and the language, and (2) predicting the reward / value of state $s$ given the goal state of the target task.
As such, we propose a neural network architecture that consists of two modules: (1) Target Goal Predictor, and (2) Reward / Value Network (see Figure~\ref{fig:nn-arch}).
This decomposition allows for additional supervision of the target goal predictor, using the ground-truth goal state for the target task.

\subsubsection{Target Goal Predictor}

Given a sequence of images representing a source demonstration ($\tau_{src}$), and a natural language description of the difference between the source and the target task ($L$),
the target goal predictor module is trained to predict the goal state of the target task ($g_{tgt}$).

\paragraph{Demonstration Encoder.} 
Each image in the source demonstration is first passed through a convolutional neural network to obtain a $D_1$-dimensional vector representation, where $D_1$ is a hyperparameter tuned using the validation set. 
The resulting sequence of vectors is padded to the maximum demonstration length ($T_{max}$) in the training data, and the vectors are then concatenated to obtain a single $T_{max}\cdot D_1$-dimensional vector.
\footnote{We also experimented with an LSTM and a transformer to encode the sequence of vectors, but obtained significantly worse performance compared to a simple concatenation. A possible explanation for this behavior is that encoding the frames into a single vector independently of the language may make it harder to associate information in language with that in individual frames, suggesting that cross-attention between language and the frames might be required. Our preliminary experiments with attention-based models did not work well, but a more thorough anaysis is needed.}


\paragraph{Language Encoder.} The natural language description is first converted into a one-hot representation using a vocabulary (see Sections~\ref{sec:lang} for details about the vocabulary), which is then passed through a two-layer LSTM module to obtain a vector representation of the description. The hidden size of the LSTM is set to $D_2$, which is tuned using the validation set.

\paragraph{Target Goal Decoder.} The vector representations of the source demonstration and the natural language description are concatenated, and the resulting vector is passed through a deconvolution network to obtain an image representation $\hat{g}$ of the target goal state.


\subsubsection{Reward / Value Network}

The reward or value network takes the predicted target goal state $\hat{g}$ and another state $s$ from the target task as inputs, and is trained to predict the reward or the value respectively of state $s$ under the target reward function.
The predicted goal state $\hat{g}$ and the state $s$ are encoded using the same CNN encoder (i.e. shared weights) used for encoding the demonstration frames in the target goal predictor, to obtain $D_1$-dimensional vector representations.
The reward or value of state $s$ is computed as the cosine similarity between the vector representations of $\hat{g}$ and the state $s$. We represent the ground-truth reward or value as $f(s)$, while the network prediction as $\hat{f}(s)$.



\subsection{Training}

To train the model, we assume access to a dataset $\mathcal{D} = \{(\tau^{i}_{src}, L^{i}, g^{i}_{tgt})\}_{i=1}^{N}$.
Using the goal state for the $i^{th}$ target task, the reward function and the optimal value function for the target task can be computed, which is used to supervise the model training as described below. 

\subsubsection{Training Objectives}

\paragraph{Mean Squared Error.}
Since we want the model to regress to the true reward or value $f(s)$ for states $s \in S^{i}_{tgt}$, a natural choice for the loss function is the mean squared error (MSE),
$\mathcal{L}_{\text{f}} = \frac{1}{N} \sum_{i=1}^{N} \sum_{s \in S^{i}_{tgt}} (f(s) - \hat{f}(s))^2$.




\paragraph{Target Goal Supervision.}
Further, we additionally supervise the Target Goal Predictor using the true goal state $g^{i}_{tgt}$ for the $i^{th}$ target task, using an MSE loss,
$\mathcal{L}_{\text{goal}} = \frac{1}{N} \sum_{i=1}^{N} (g^{i}_{tgt} - \hat{g}^{i}_{tgt})^2$.



Thus, the overall training loss is given by
$ \mathcal{L} = \mathcal{L}_{\text{f}} + \lambda \mathcal{L}_{\text{goal}} $,
where $\lambda$ is an hyperparameter tuned using a validation set.

\subsubsection{Optimization}

For training the model, a datapoint $(\tau^{i}_{src}, L^{i}, g^{i}_{tgt})$ is sampled uniformly at random from $\mathcal{D}$. When predicting the value function, a target state $s$ is sampled uniformly at random from $S^{i}_{tgt}$ at each step of the optimization process. When predicting the reward function, the goal state $g^{i}_{tgt}$ is sampled with 50\% probability, while a non-goal state is sampled uniformly at random otherwise. This is required because the reward function is sparse.
We use an Adam optimizer \cite{kingma2014adam} to train the network end-to-end for 50 epochs, with weights initialized randomly  according to \textcite{glorot2010understanding}.
A validation set is used to tune hyperparameters via random search.

\section{Environment and Dataset}
\label{sec:data}

In this section, we describe the environment we use in our experiments. While the framework described above is fairly general, in this work, we focus on a simpler setting that is more amenable to analysis. Specifically, we assume discrete state and action spaces $S$ and $A$, and deterministic transitions, i.e., $P(s, a, s') \in \{0, 1\}, \forall (s, a, s') \in S \times A \times S$.

\subsection{The Organizer Environment}
\label{sec:env}

We propose the Organizer Environment, which consists of an organizer with 3 shelves. 
There are 8 distinct objects, and each object can take one of 3 colors (red, blue, or green), giving a total of 24 distinct colored objects (see Figure~\ref{fig:objects} in the appendix).
Objects can be placed in each shelf, either to the left or the right, resulting in a total of 6 distinct positions, 
$\texttt{POSITIONS} =$ $\{\texttt{Top-Left},  \texttt{Top-Right}, \texttt{Middle-Left},  \texttt{Middle-Right}, \texttt{Bottom-Left},  \texttt{Bottom-Right}\}$.

Objects can be placed in different configurations to create different states. In our experiments, we use tasks with 2 or 3 objects. The total number of states with 2 or 3 objects (i.e. $| \bigcup_{T \in \mathcal{T}} S_{T} |$) is 285,120.
The action space $A$ is common across all tasks, and consists of 30 move actions, $\texttt{MOVE}(p_{i}, p_{j}), p_{i}, p_{j} \in \texttt{POSITIONS}, p_{i} \ne p_{j}$. Finally, there is a $\texttt{STOP}$ action that indicates the termination of an episode.


\subsection{Language Data}
\label{sec:lang}

In this work, we experiment with 6 types of adaptations: (1) moving the same object in the source and target tasks, but to different final positions; (2) moving a different object in the source and target tasks, but to the same final position; (3) moving two objects in the source and target tasks, with their final positions swapped in the target task; (4) deleting a step from the source task; (5) inserting a step to the source task; and (6) modifying a step in the source task.
Examples for each of these adaptations are shown in the appendix (Table~\ref{tbl:adaptations}).

For each pair of source and target tasks in the dataset, we need a linguistic description of the difference between the tasks.
We start by generating linguistic descriptions using a set of templates, such as, ``Move $\texttt{obj1}$ instead of $\texttt{obj2}$ to the same position.''
We ensure that for most of these templates, the target task cannot be inferred from the description alone, and thus, the model must use both the demonstration of the source task and the linguistic description to infer the goal for the target task.

Next, we collected natural language for a subset of these synthetic (i.e. template-generated) descriptions using Amazon Mechanical Turk (AMT).
Workers were provided with the template-generated descriptions, and were asked to paraphrase these descriptions.
Importantly, our initial data-collection experiments suggested that providing the workers with the task images resulted in inferior descriptions, wherein, many descriptions would describe the target task completely, instead of how it differs from the source task.
As such, the workers were first shown some examples of source and target tasks to familiarize them with the domain, and were then only provided with template-generated descriptions, without the images for the source and target tasks, to obtain the paraphrases. See Section~\ref{sec:amt} in the appendix for more details about the data collection process.


We applied a basic filtering process to remove clearly bad descriptions, such as those with 2 words or less, and those that were identical to the given paraphrases. We did not make any edits to the descriptions, like correcting for grammatical or spelling errors.
Some examples of template-generated and natural language descriptions obtained using AMT are shown in Table~\ref{tbl:syn-nat-examples}.

It can be observed that while the first four rows in the table are valid paraphrases, the remaining three paraphrases could be ambiguous depending on the source and target tasks.
For instance, in row 5, the target task involves an \emph{extra} step after the first step, while the natural language paraphrase could be interpreted as \emph{modifying} the second step.
In row 6, the natural language description is not a valid paraphrase, while in row 7, the paraphrase is difficult to comprehend.
We manually analysed a small subset of the collected paraphrases, and found that about 15\% of the annotations were ambiguous / non-informative.
Some of this noise could be addressed by modifying the data-collection pipeline, for example, by providing more information about the source and target tasks to humans, and filtering out non-informative / difficult to comprehend paraphrases.

\begin{table}[]
\caption{Examples of template-generated and natural language descriptions collected using AMT.}
\small
\begin{tabular}{l p{0.435\textwidth}p{0.435\textwidth}}
\hline
  & \Tstrut {\textbf{Template}}                                              & {\textbf{Natural language paraphrase}}\Bstrut                    \\
\hline
1.
& \Tstrut Move the cylinder to middle left.
& Place the cylinder in the middle left
\Bstrut \\ \hline
2.
& \Tstrut Move the red tall block to the final position of green long block.
& Place the red tall block in the green longs blocks final position
\Bstrut \\ \hline
3.
& \Tstrut Skip the third step.
& do not do the third step
\Bstrut \\ \hline
4.
& \Tstrut In the first step, move the green tall cylinder from bottom right to bottom left.
& for the first step, put the green tall cylinder in the bottom left position
\Bstrut \\ \hline
5.
& \Tstrut Move blue tall cylinder from bottom left to middle left after the first step.
& For the second step move the blue tall cylinder to the middle left
\Bstrut \\ \hline
6.
& \Tstrut Move the blue cube to the final position of blue tall cylinder.
& Swap the blue cube with the red cube on bottom shelf.
\Bstrut \\ \hline
7.
& \Tstrut Move blue tall block from top right to bottom left after the second step.
& Move blue tall square from upper option to base left after the subsequent advance.
\Bstrut \\ \hline
\end{tabular}
\label{tbl:syn-nat-examples}
\end{table}

A vocabulary was created using the training split of the synthetic and natural language descriptions, discarding words that occurred fewer than 10 times in the corpus. 
While encoding a description using the resulting vocabulary, out-of-vocabulary tokens were represented using the \texttt{<unk>} symbol.




\section{Experiments}
\label{sec:expts}

\subsection{Details about the setup}


\paragraph{Dataset.} For each adaptation, 6,600 pairs of source and target tasks were generated along with the template-based descriptions.
Of these, 600 templates were used for each adaptation to collect natural language descriptions using Amazon Mechanical Turk.
Thus, our dataset consisted of 6,600 examples in total for each adaptation, of which 6,000 examples consisted of synthetic language, and 600 consisted of natural language.
Of the 6,000 synthetic examples per adaptation, 5,000 were used for training, 500 for validation, and the remaining 500 for testing.
Similarly, of the 600 natural language examples per adaptation, 500 were used for training, 50 for validation, and 50 for testing.
This gave us a training dataset with 33,000 examples, and validation and test datasets with 3,300 examples each, across all adaptation types.

\paragraph{Evaluation Metrics.} For each experiment, the trained model predicts the reward or value of the given state $s$. 
When using the value function, the trained network is used to predict the values for all states $s \in S_{tgt}$.
When using the reward function, the trained network is used to predict the rewards for all states $s \in S_{tgt}$, from which the optimal value function is computed using dynamic programming.
In both cases, if the state with the maximum value matches the goal state for the target task, $g_{tgt}$, the task is deemed to be successful.
We train the models using the entire training set (i.e. both synthetic and natural language examples across all adaptations), and report the percentage of successfully completed target tasks for both synthetic and natural language descriptions. For each experiment, we tune the hyperparameters on the validation set, and report the success rate on the test set corresponding to the setting yielding the maximum success rate on the validation set.

\begin{table}[]
\caption{Success rates of different models}
\centering
\small
\begin{tabular}{llrr}
\hline
\multirow{2}{*}{} & \multicolumn{1}{c}{\multirow{2}{*}{\textbf{Experiment}}} & \multicolumn{2}{c}{\textbf{Success rate (\%)}}                                      \\ \cline{3-4} 
                  & \multicolumn{1}{c}{}                                & \multicolumn{1}{c}{\textbf{Synthetic}} & \multicolumn{1}{c}{\textbf{Natural}} \\ \hline
1. & LARVA; reward prediction                     & 97.8       & 75.7        \\ \hline
2. & LARVA; value prediction                      & 97.7       & 73.3        \\ \hline
\hline
3. & LARVA; no target goal supervision            & 20.0       &  2.7        \\ \hline
4. & LARVA; no language                           & 20.7       & 22.3        \\ \hline
5. & LARVA; no source demonstration               &  4.2       &  3.3        \\ \hline
6. & NN without decomposition                     &  1.8       &  1.0        \\ \hline
\hline
7. & LARVA: Compostionality -- red box            & 87.6       & 62.4        \\ \hline
8. & LARVA: Compostionality -- blue cylinder      & 89.4       & 65.9        \\ \hline
\end{tabular}
\label{tbl:results}
\end{table}

\subsection{Results}

In this section, we describe the performance of our full model, along with various ablations. Our results are summarized in Table~\ref{tbl:results}.

First, we evaluate our full LARVA model, with both reward and value predictions (rows 1 and 2 in Table~\ref{tbl:results}). 
In both cases, the models result in successful completion of the target task more than 97\% of the time with synthetic language, and more than 73\% of the time with natural language. 
The drop in performance when using natural language can partly be attributed to the 15\% of paraphrases that are potentially ambiguous or uninformative, as discussed in Section~\ref{sec:lang}, while the remaining performance gap is likely because natural language has more variation than synthetic language.
Better data collection and more complex models could be explored to bridge this gap further. Our experiments to analyze the impact of the quantity of data suggests that increasing the amount of synthetic or natural language data is unlikely to provide a significant improvement on the natural language test set (see Section~\ref{sec:add-expts} in the appendix).

The similar performance when predicting rewards and values is not unexpected---we observed in our experiments that training the target goal prediction module was more challenging than training the the reward or value networks. Since the target goal prediction module is identical for both reward and value predictions, the performance in both cases is upper-bounded by the performance of the target goal prediction module. For domains with complex dynamics, reward and value prediction might result in significantly different sucess rates.

Next, we discuss ablations of our model. We present the results only with value prediction, since as noted, both reward and value predictions perform similarly.
    
\begin{enumerate}[leftmargin=12pt]
    \item To study the effect of target goal supervision for training the target goal predictor, we remove $\mathcal{L}_{goal}$, optimizing the network using the ground-truth values only. Row 3 in Table~\ref{tbl:results} shows that this drastically degrades performance, confirming the efficacy of target goal supervision.
    \item To ensure that most tasks require using information from both the source demonstration and the natural language description, we run unimodal baselines, wherein the network is provided with only the source demonstration (row 4) or only the language (row 5). As expected, both the settings result in a substantial drop in performance. Interestingly, using only the source demonstration results in over 20\% successful completions. This is because given the set of adaptations, the source demonstration constrains the space of target configurations more effectively (e.g. if the source task consists of three steps, the target task must contain at least two of those steps, since source and target tasks differ by only one step for all adaptations).
    \item Finally, we experiment with an alternate neural network architecture, that does not decompose the learning problem into target goal prediction and value prediction. The source demonstration, the language, and the target state $s$ are all encoded independently, and concatenated, from which the value for state $s$ is predicted. Row 6 shows that the resulting model achieves a very low success on target tasks, demonstrating the importance of decomposing the learning problem as in LARVA.
\end{enumerate}


In the experiments so far, the data were randomly partitioned into training, validation, and test splits. 
However, a key aspect of using language is the ability to \emph{compose} concepts.
For instance, humans can learn concepts like ``blue box'' and ``red cylinder'' from independent examples, and can recognize a ``blue cylinder'' by composing these concepts without ever having seen examples of the new concept.

To evaluate whether our proposed model can exhibit the ability to compose concepts seen during training, we create 2 new splits of our data---in both the splits, the training data consists of all datapoints that do not contain any blue cylinders or red boxes. 
In the first split, the validation set consists of all datapoints containing blue cylinders, while the test set consists of all datapoints containing red boxes. In the second split, the validation and test sets are swapped.\footnote{Datapoints containing both a blue cylinder and a red box are discarded.}

We train LARVA on these new splits (using value prediction), and report the success rate on the test set in Table~\ref{tbl:results}, rows 7 and 8.
As can be observed, our model is able to successfully complete a large fraction of the target tasks, by composing concepts seen during training, however, there is room for further improvement by using richer models.




\section{Conclusions and Future Work}
\label{sec:discussion}

\paragraph{Conclusions.} We proposed a novel framework which allows teaching agents using a combination of demonstrations and language.
Given a demonstration of a source task, and a natural language description of the difference between the source task and a target task, we introduce Language-Aided Reward and Value Adaptation (LARVA), a model that can perform the target task in a zero-shot setting.
We experimented with a diverse set of adaptations on a simple discrete environment, and show that the model is able to complete nearly all target tasks successfully when using synthetic language, while more than 70\% of the target tasks when using free-form natural language.
A key component of LARVA, as demonstrated by the ablation experiments, is decomposing the full problem into two subproblems (target goal prediction and reward / value prediction), which allows for intermediate supervision.

\paragraph{Limitations and Future work.} 
First, our experimental evaluation involved a fairly simple setup. While a simple domain allows for faster experimentation and better analysis, richer domains with more visual diversity, complex dynamics, and continuous states and actions ought to provide a more thorough analysis in future work. 
Similarly, a relatively simple family of tasks was considered in this work. Generalization to more complex families of tasks needs further experimentation.
It is worth noting that our general framework is applicable to all these variations.
Second, our approach requires about 30,000 pairs of source and target tasks, along with natural language descriptions to learn the model. While this is comparable to related approaches (e.g. \textcite{stepputtis2020language} use 40,000 training examples for instruction-following), we believe that on more realistic tasks, using pretrained vision and language encoders could significantly reduce the data requirements. Further, training the model is a one-time process, that can be performed before deployment; after deployment, the system can be provided with demonstrations and descriptions to perform new tasks without any further training.
Finally, when using natural language, there is a significant performance gap from the template-generated language. A data collection pipeline with better noise-filtering and richer language models (e.g. ones that use attention) could help bridge this gap.

\newpage

\nocite{zhu2020robosuite}
\printbibliography

@article{goyal2019using,
  title={Using natural language for reward shaping in reinforcement learning},
  author={Goyal, Prasoon and Niekum, Scott and Mooney, Raymond J},
  journal={arXiv preprint arXiv:1903.02020},
  year={2019}
}

@article{goyal2020pixl2r,
  title={PixL2R: Guiding Reinforcement Learning Using Natural Language by Mapping Pixels to Rewards},
  author={Goyal, Prasoon and Niekum, Scott and Mooney, Raymond J},
  journal={arXiv preprint arXiv:2007.15543},
  year={2020}
}

@inproceedings{anderson2018vision,
  title={Vision-and-language navigation: Interpreting visually-grounded navigation instructions in real environments},
  author={Anderson, Peter and Wu, Qi and Teney, Damien and Bruce, Jake and Johnson, Mark and S{\"u}nderhauf, Niko and Reid, Ian and Gould, Stephen and Van Den Hengel, Anton},
  booktitle={Proceedings of the IEEE Conference on Computer Vision and Pattern Recognition},
  pages={3674--3683},
  year={2018}
}

@inproceedings{wang2019reinforced,
  title={Reinforced cross-modal matching and self-supervised imitation learning for vision-language navigation},
  author={Wang, Xin and Huang, Qiuyuan and Celikyilmaz, Asli and Gao, Jianfeng and Shen, Dinghan and Wang, Yuan-Fang and Wang, William Yang and Zhang, Lei},
  booktitle={Proceedings of the IEEE/CVF Conference on Computer Vision and Pattern Recognition},
  pages={6629--6638},
  year={2019}
}

@article{fried2018speaker,
  title={Speaker-follower models for vision-and-language navigation},
  author={Fried, Daniel and Hu, Ronghang and Cirik, Volkan and Rohrbach, Anna and Andreas, Jacob and Morency, Louis-Philippe and Berg-Kirkpatrick, Taylor and Saenko, Kate and Klein, Dan and Darrell, Trevor},
  journal={arXiv preprint arXiv:1806.02724},
  year={2018}
}

@inproceedings{tellex2011understanding,
  title={Understanding natural language commands for robotic navigation and mobile manipulation},
  author={Tellex, Stefanie and Kollar, Thomas and Dickerson, Steven and Walter, Matthew and Banerjee, Ashis and Teller, Seth and Roy, Nicholas},
  booktitle={Proceedings of the AAAI Conference on Artificial Intelligence},
  volume={25},
  number={1},
  year={2011}
}

@inproceedings{hemachandra2015learning,
  title={Learning models for following natural language directions in unknown environments},
  author={Hemachandra, Sachithra and Duvallet, Felix and Howard, Thomas M and Roy, Nicholas and Stentz, Anthony and Walter, Matthew R},
  booktitle={2015 IEEE International Conference on Robotics and Automation (ICRA)},
  pages={5608--5615},
  year={2015},
  organization={IEEE}
}

@inproceedings{arkin2017contextual,
  title={Contextual awareness: Understanding monologic natural language instructions for autonomous robots},
  author={Arkin, Jacob and Walter, Matthew R and Boteanu, Adrian and Napoli, Michael E and Biggie, Harel and Kress-Gazit, Hadas and Howard, Thomas M},
  booktitle={2017 26th IEEE International Symposium on Robot and Human Interactive Communication (RO-MAN)},
  pages={502--509},
  year={2017},
  organization={IEEE}
}

@inproceedings{shridhar2020alfred,
  title={Alfred: A benchmark for interpreting grounded instructions for everyday tasks},
  author={Shridhar, Mohit and Thomason, Jesse and Gordon, Daniel and Bisk, Yonatan and Han, Winson and Mottaghi, Roozbeh and Zettlemoyer, Luke and Fox, Dieter},
  booktitle={Proceedings of the IEEE/CVF conference on computer vision and pattern recognition},
  pages={10740--10749},
  year={2020}
}

@article{stepputtis2020language,
  title={Language-Conditioned Imitation Learning for Robot Manipulation Tasks},
  author={Stepputtis, Simon and Campbell, Joseph and Phielipp, Mariano and Lee, Stefan and Baral, Chitta and Amor, Heni Ben},
  journal={arXiv preprint arXiv:2010.12083},
  year={2020}
}

@incollection{sung2018robobarista,
  title={Robobarista: Object part based transfer of manipulation trajectories from crowd-sourcing in 3d pointclouds},
  author={Sung, Jaeyong and Jin, Seok Hyun and Saxena, Ashutosh},
  booktitle={Robotics Research},
  pages={701--720},
  year={2018},
  publisher={Springer}
}

@article{andreas2017learning,
  title={Learning with latent language},
  author={Andreas, Jacob and Klein, Dan and Levine, Sergey},
  journal={North American Chapter of the Association for Computational Linguistics (NAACL)},
  year={2018}
}

@article{branavan2012learning,
  title={Learning to win by reading manuals in a Monte-Carlo framework},
  author={Branavan, SRK and Silver, David and Barzilay, Regina},
  journal={Journal of Artificial Intelligence Research},
  volume={43},
  year={2012}
}

@article{narasimhan2018grounding,
  title={Grounding language for transfer in deep reinforcement learning},
  author={Narasimhan, Karthik and Barzilay, Regina and Jaakkola, Tommi},
  journal={Journal of Artificial Intelligence Research},
  volume={63},
  pages={849--874},
  year={2018}
}

@article{kaplan2017beating,
  title={Beating atari with natural language guided reinforcement learning},
  author={Kaplan, Russell and Sauer, Christopher and Sosa, Alexander},
  journal={arXiv preprint arXiv:1704.05539},
  year={2017}
}

@article{wang2021grounding,
  title={Grounding Language to Entities and Dynamics for Generalization in Reinforcement Learning},
  author={Wang, HJ and Narasimhan, Karthik},
  journal={arXiv preprint arXiv:2101.07393},
  year={2021}
}

@article{kingma2014adam,
  title={Adam: A method for stochastic optimization},
  author={Kingma, Diederik P and Ba, Jimmy},
  journal={International Conference for Learning Representations (ICLR)},
  year={2015}
}

@article{argall2009survey,
  title={A survey of robot learning from demonstration},
  author={Argall, Brenna D and Chernova, Sonia and Veloso, Manuela and Browning, Brett},
  journal={Robotics and autonomous systems},
  volume={57},
  number={5},
  pages={469--483},
  year={2009},
  publisher={Elsevier}
}

@inproceedings{abbeel2004apprenticeship,
  title={Apprenticeship learning via inverse reinforcement learning},
  author={Abbeel, Pieter and Ng, Andrew Y},
  booktitle={Proceedings of the twenty-first international conference on Machine learning},
  pages={1},
  year={2004}
}

@inproceedings{ramachandran2007bayesian,
  title={Bayesian Inverse Reinforcement Learning.},
  author={Ramachandran, Deepak and Amir, Eyal},
  booktitle={IJCAI},
  volume={7},
  pages={2586--2591},
  year={2007}
}

@inproceedings{ziebart2008maximum,
  title={Maximum entropy inverse reinforcement learning.},
  author={Ziebart, Brian D and Maas, Andrew L and Bagnell, J Andrew and Dey, Anind K},
  booktitle={Aaai},
  volume={8},
  pages={1433--1438},
  year={2008},
  organization={Chicago, IL, USA}
}

@inproceedings{finn2016guided,
  title={Guided cost learning: Deep inverse optimal control via policy optimization},
  author={Finn, Chelsea and Levine, Sergey and Abbeel, Pieter},
  booktitle={International conference on machine learning},
  pages={49--58},
  year={2016},
  organization={PMLR}
}

@article{fu2017learning,
  title={Learning robust rewards with adversarial inverse reinforcement learning},
  author={Fu, Justin and Luo, Katie and Levine, Sergey},
  journal={arXiv preprint arXiv:1710.11248},
  year={2017}
}

@article{ho2016generative,
  title={Generative adversarial imitation learning},
  author={Ho, Jonathan and Ermon, Stefano},
  journal={arXiv preprint arXiv:1606.03476},
  year={2016}
}

@techreport{pomerleau1989alvinn,
  title={Alvinn: An autonomous land vehicle in a neural network},
  author={Pomerleau, Dean A},
  year={1989},
  institution={CARNEGIE-MELLON UNIV PITTSBURGH PA ARTIFICIAL INTELLIGENCE AND PSYCHOLOGY~…}
}

@inproceedings{ross2010efficient,
  title={Efficient reductions for imitation learning},
  author={Ross, St{\'e}phane and Bagnell, Drew},
  booktitle={Proceedings of the thirteenth international conference on artificial intelligence and statistics},
  pages={661--668},
  year={2010},
  organization={JMLR Workshop and Conference Proceedings}
}

@inproceedings{ross2011reduction,
  title={A reduction of imitation learning and structured prediction to no-regret online learning},
  author={Ross, St{\'e}phane and Gordon, Geoffrey and Bagnell, Drew},
  booktitle={Proceedings of the fourteenth international conference on artificial intelligence and statistics},
  pages={627--635},
  year={2011},
  organization={JMLR Workshop and Conference Proceedings}
}

@article{pan2009survey,
  title={A survey on transfer learning},
  author={Pan, Sinno Jialin and Yang, Qiang},
  journal={IEEE Transactions on knowledge and data engineering},
  volume={22},
  number={10},
  pages={1345--1359},
  year={2009},
  publisher={IEEE}
}

@article{misra2016tell,
  title={Tell me dave: Context-sensitive grounding of natural language to manipulation instructions},
  author={Misra, Dipendra K and Sung, Jaeyong and Lee, Kevin and Saxena, Ashutosh},
  journal={The International Journal of Robotics Research},
  volume={35},
  number={1-3},
  pages={281--300},
  year={2016},
  publisher={SAGE Publications Sage UK: London, England}
}

@article{harnad1990symbol,
  title={The symbol grounding problem},
  author={Harnad, Stevan},
  journal={Physica D: Nonlinear Phenomena},
  volume={42},
  number={1-3},
  pages={335--346},
  year={1990},
  publisher={Elsevier}
}

@inproceedings{luketina2019survey,
  title = "A Survey of Reinforcement Learning Informed by Natural Language",
  author = "Jelena Luketina and Nantas Nardelli and Gregory Farquhar and Jakob Foerster and Jacob Andreas and Edward Grefenstette and Shimon Whiteson and Tim Rocktaschel",
  year = "2019",
  booktitle = "IJCAI 2019: Proceedings of the Twenty-Eighth International Joint Conference on Artificial Intelligence",
  month = "August",
}

@article{macmahon2006walk,
  title={Walk the talk: Connecting language, knowledge, and action in route instructions},
  author={MacMahon, Matt and Stankiewicz, Brian and Kuipers, Benjamin},
  journal={Def},
  volume={2},
  number={6},
  pages={4},
  year={2006}
}

@inproceedings{vogel2010learning,
  title={Learning to follow navigational directions},
  author={Vogel, Adam and Jurafsky, Dan},
  booktitle={Proceedings of the 48th Annual Meeting of the Association for Computational Linguistics},
  pages={806--814},
  year={2010}
}

@inproceedings{chen2011learning,
  title={Learning to interpret natural language navigation instructions from observations},
  author={Chen, David and Mooney, Raymond},
  booktitle={Proceedings of the AAAI Conference on Artificial Intelligence},
  volume={25},
  number={1},
  year={2011}
}

@article{vanschoren2018meta,
  title={Meta-learning: A survey},
  author={Vanschoren, Joaquin},
  journal={arXiv preprint arXiv:1810.03548},
  year={2018}
}

@article{wang2020generalizing,
  title={Generalizing from a few examples: A survey on few-shot learning},
  author={Wang, Yaqing and Yao, Quanming and Kwok, James T and Ni, Lionel M},
  journal={ACM Computing Surveys (CSUR)},
  volume={53},
  number={3},
  pages={1--34},
  year={2020},
  publisher={ACM New York, NY, USA}
}

@inproceedings{glorot2010understanding,
  title={Understanding the difficulty of training deep feedforward neural networks},
  author={Glorot, Xavier and Bengio, Yoshua},
  booktitle={Proceedings of the thirteenth international conference on artificial intelligence and statistics},
  pages={249--256},
  year={2010},
  organization={JMLR Workshop and Conference Proceedings}
}

@article{zhu2020robosuite,
  title={robosuite: A modular simulation framework and benchmark for robot learning},
  author={Zhu, Yuke and Wong, Josiah and Mandlekar, Ajay and Martin-Martin, Roberto},
  journal={arXiv preprint arXiv:2009.12293},
  year={2020}
}

@article{kroemer2019review,
  title={A review of robot learning for manipulation: Challenges, representations, and algorithms},
  author={Kroemer, Oliver and Niekum, Scott and Konidaris, George},
  journal={arXiv preprint arXiv:1907.03146},
  year={2019}
}



\appendix

\section{Data Collection using Amazon Mechanical Turk}
\label{sec:amt}

\begin{figure}[h]
    \centering
    \includegraphics[width=0.45\textwidth]{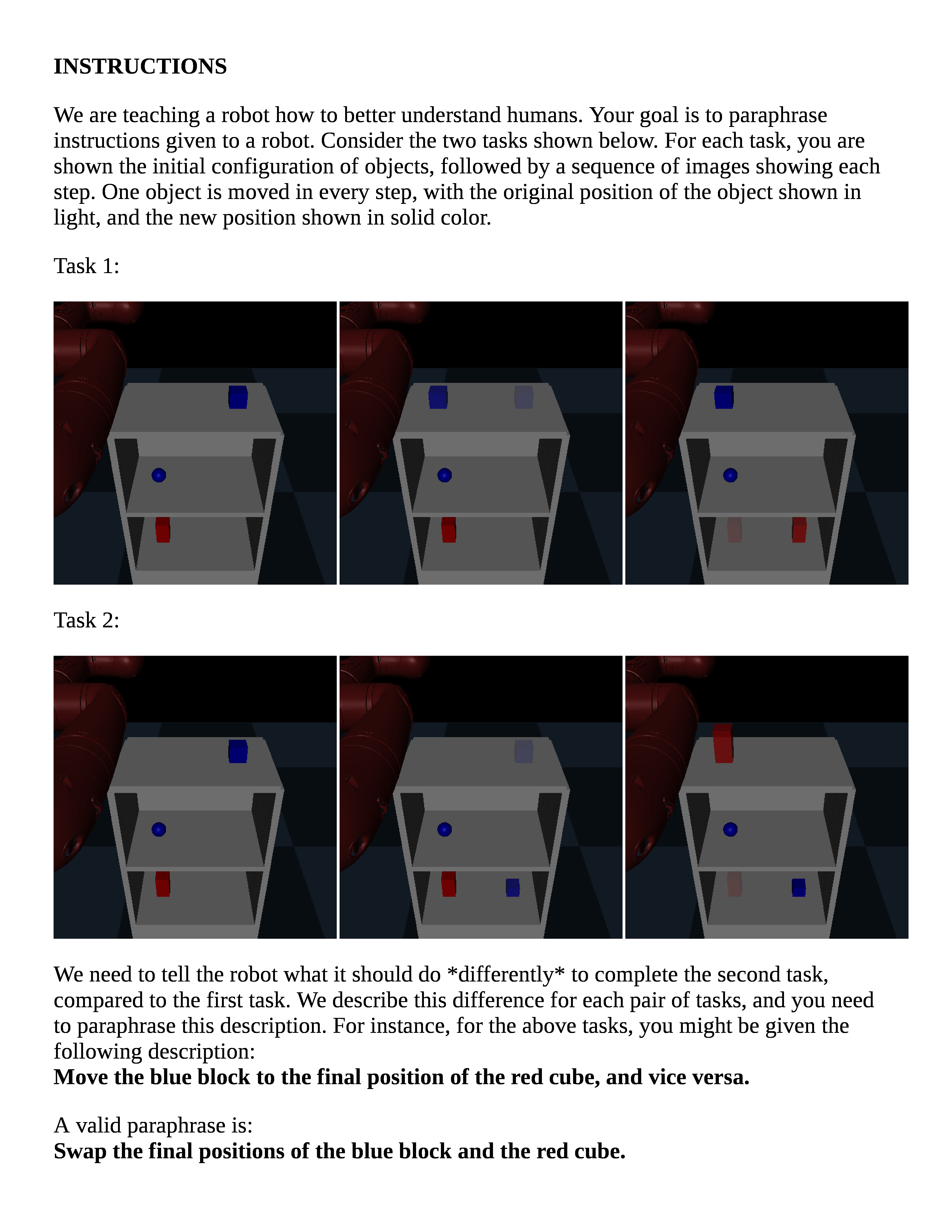}
    \includegraphics[width=0.45\textwidth]{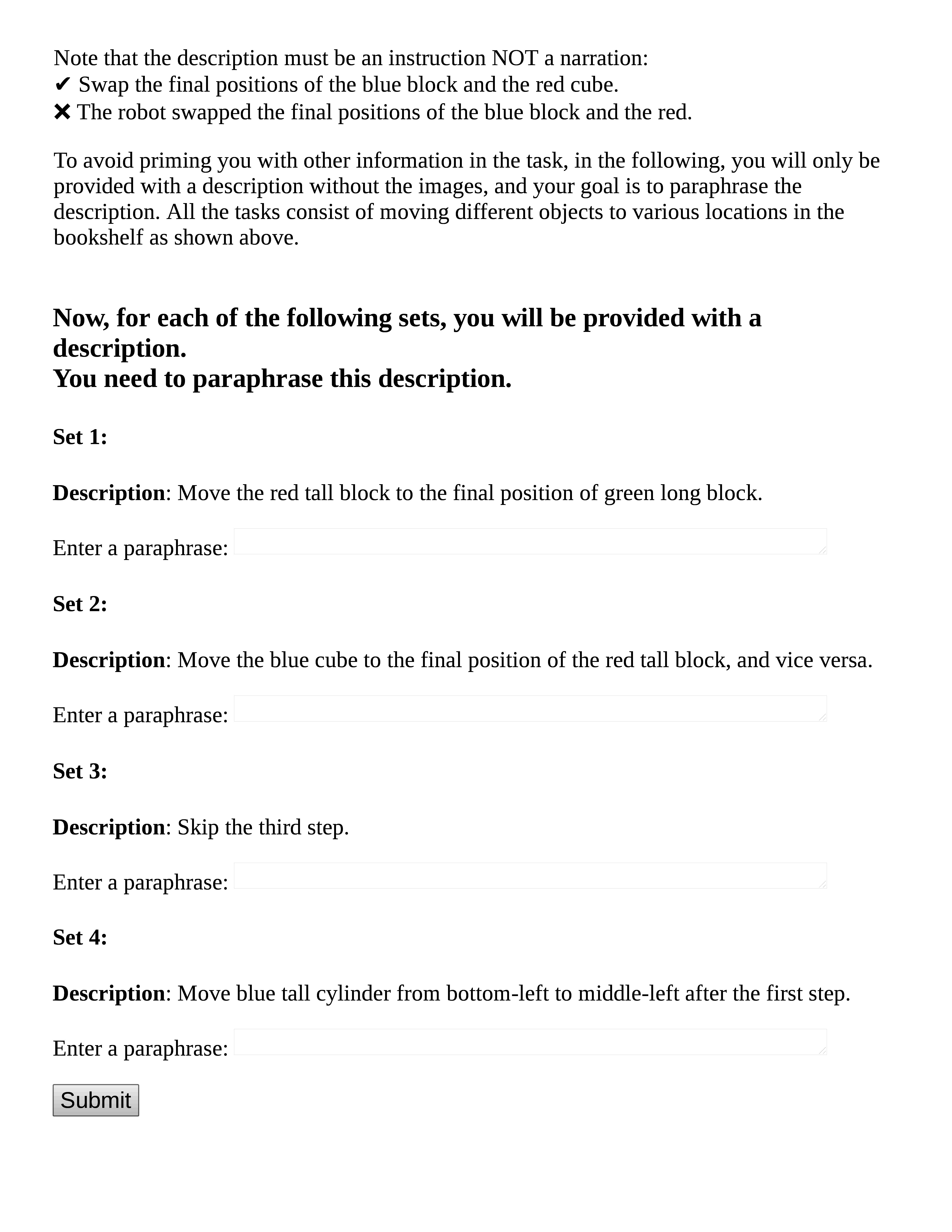}
    \caption{Interface for collecting paraphrases using Amazon Mechanical Turk}
    \label{fig:amt}
\end{figure}

The interface used for collecting natural language descriptions using Amazon Mechanical Turk is shown in Figure~\ref{fig:amt}.
The total amount of money spent on data collection was \$985, with an average pay of \$6.78 per hour.

\section{Details about the Domain}

\def \imgscale{0.08}

\begin{table}[]
\small
\caption{Types of adaptations used in our experiments. For each type, an example of source and target tasks is shown.}
\begin{tabular}{lll}
\hline
1.					          & \textbf{Same object, different place location}\Tstrut             &                                                  \\
                              & \includegraphics[scale=\imgscale]{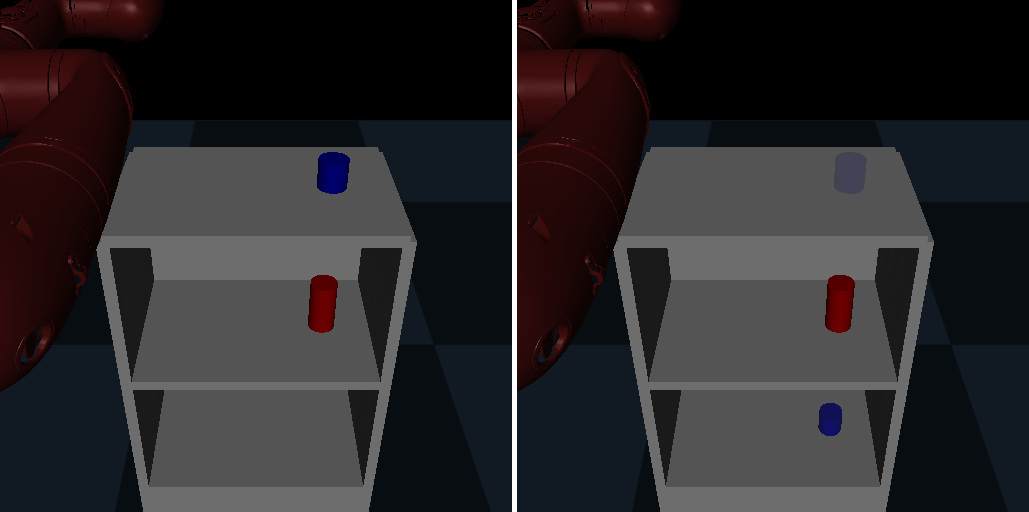} & \includegraphics[scale=\imgscale]{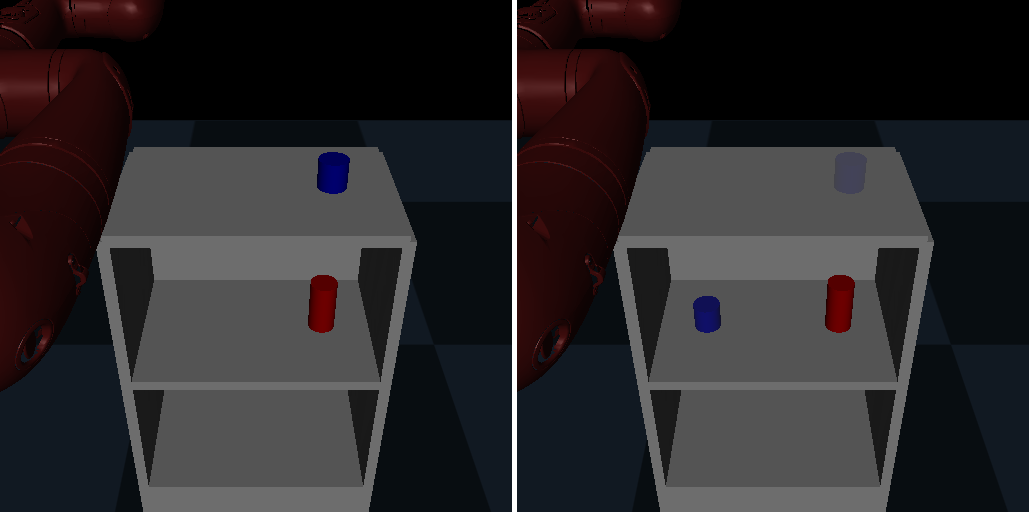} \\ 
                              &	Example Source Task\Bstrut 							& Example Target Task								\\ \hline
2.					          & \textbf{Different object, same place location}\Tstrut           &                                                  \\
                              & \includegraphics[scale=\imgscale]{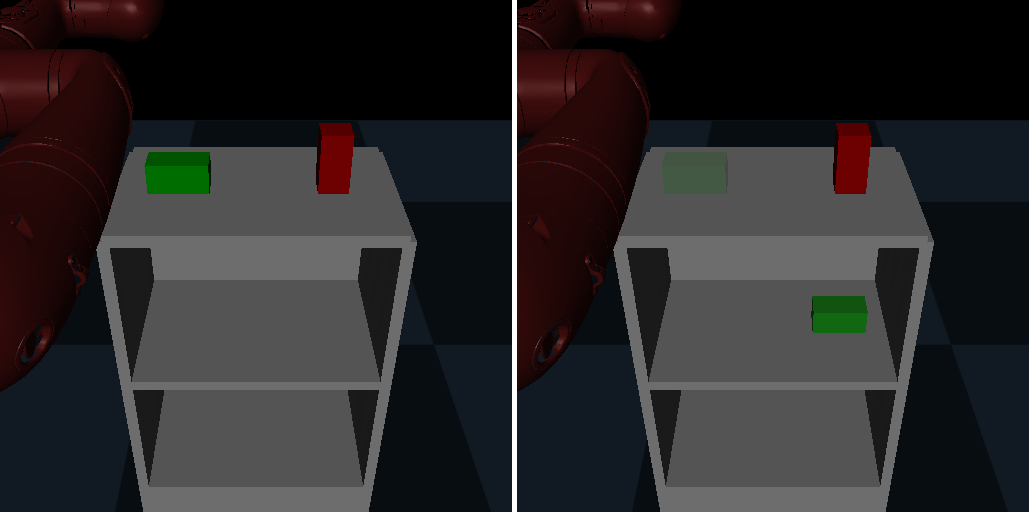} & \includegraphics[scale=\imgscale]{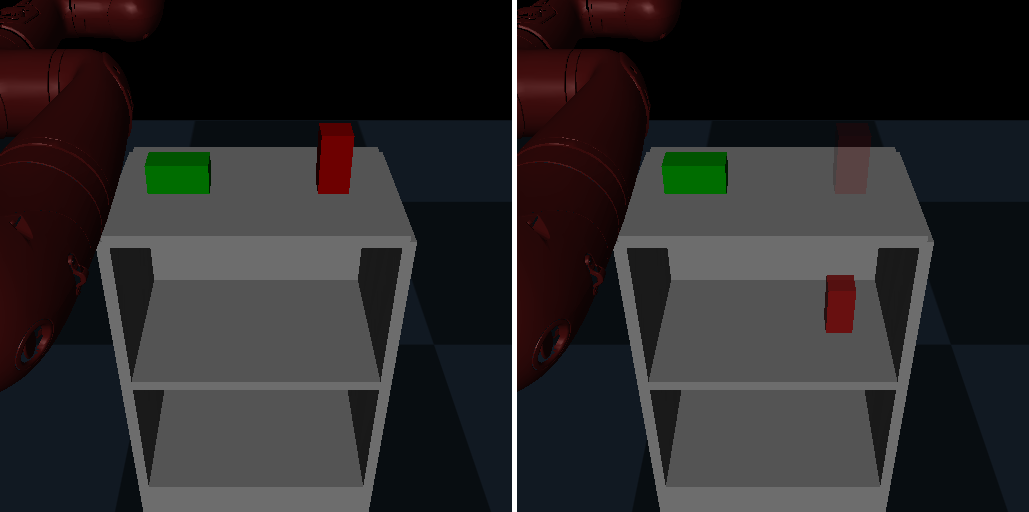} \\ 
                              &	Example Source Task\Bstrut 							& Example Target Task								\\ \hline
3.					          & \textbf{Move two objects, with swapped final locations}\Tstrut      &                                                  \\
                              & \includegraphics[scale=\imgscale]{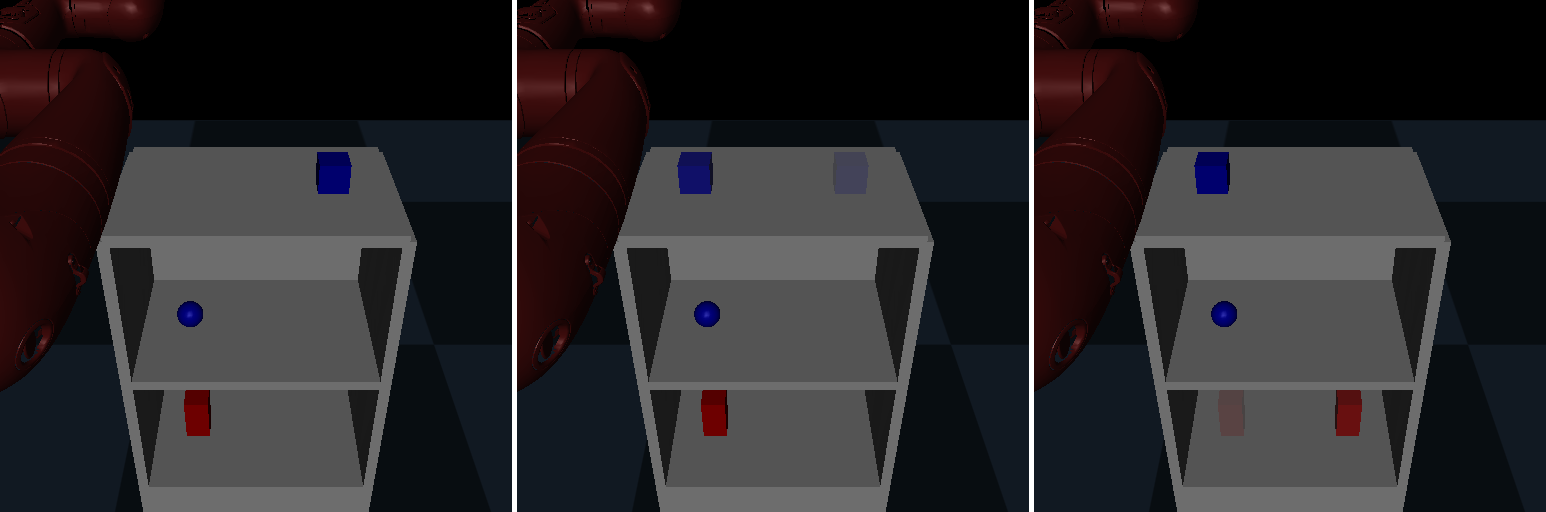} & \includegraphics[scale=\imgscale]{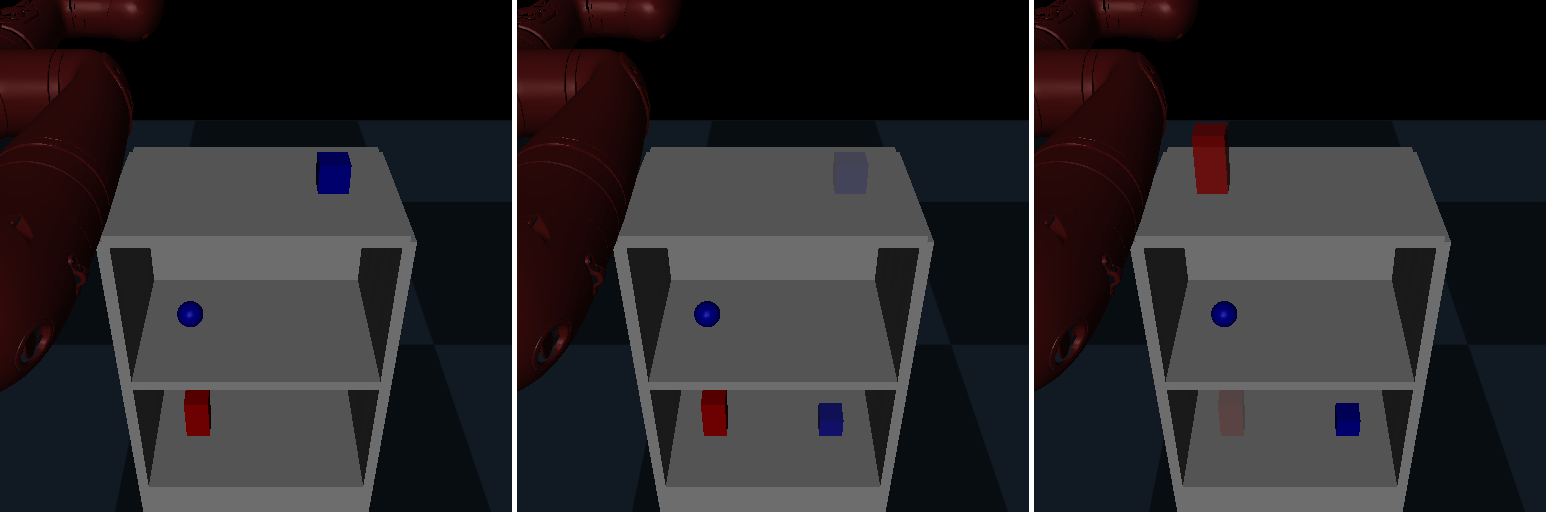} \\ 
                              &	Example Source Task\Bstrut 							& Example Target Task								\\ \hline
4.					          & \textbf{Delete a step}\Tstrut                                   &                                                  \\
                              & \includegraphics[scale=\imgscale]{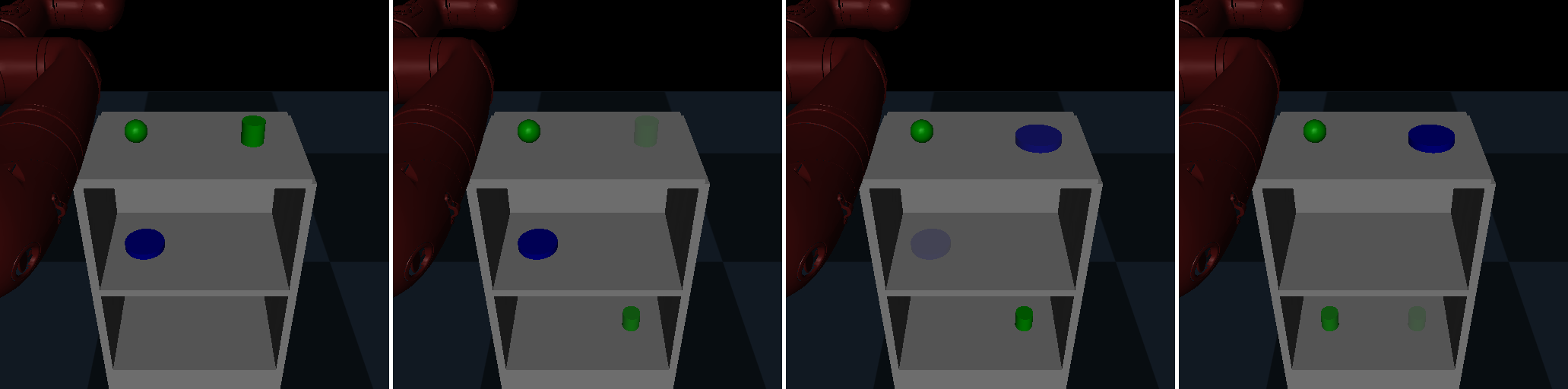} & \includegraphics[scale=\imgscale]{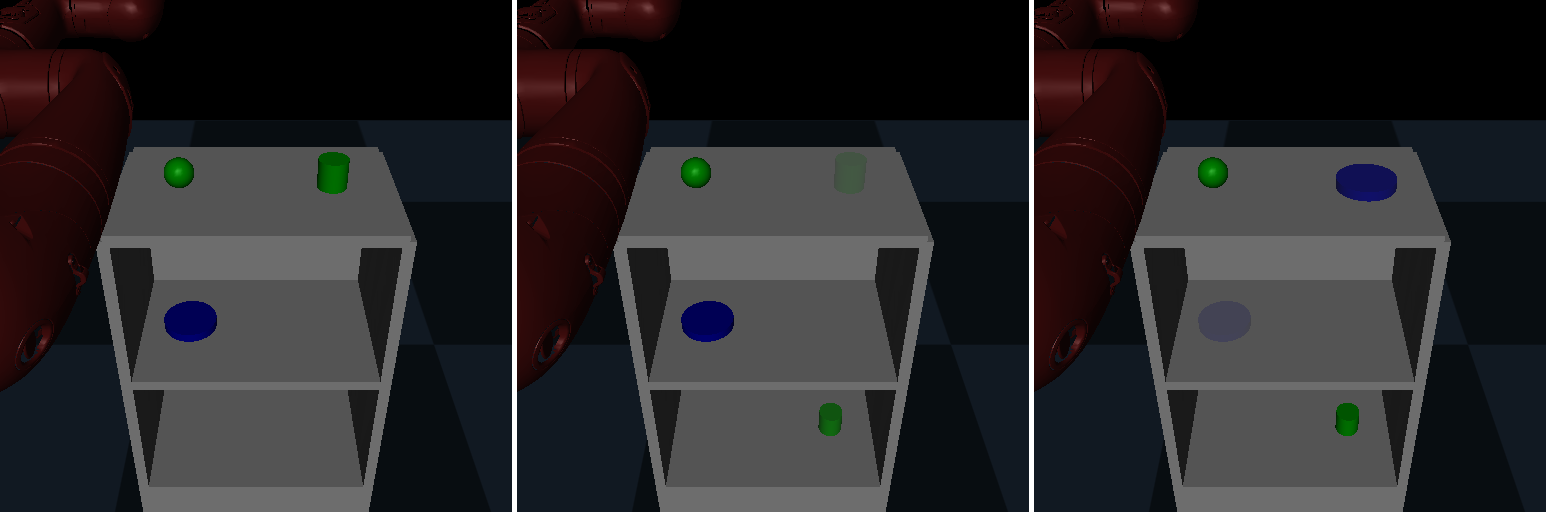} \\ 
                              &	Example Source Task\Bstrut 							& Example Target Task								\\ \hline
5.					          & \textbf{Insert a step}\Tstrut                                   &                                                  \\
                              & \includegraphics[scale=\imgscale]{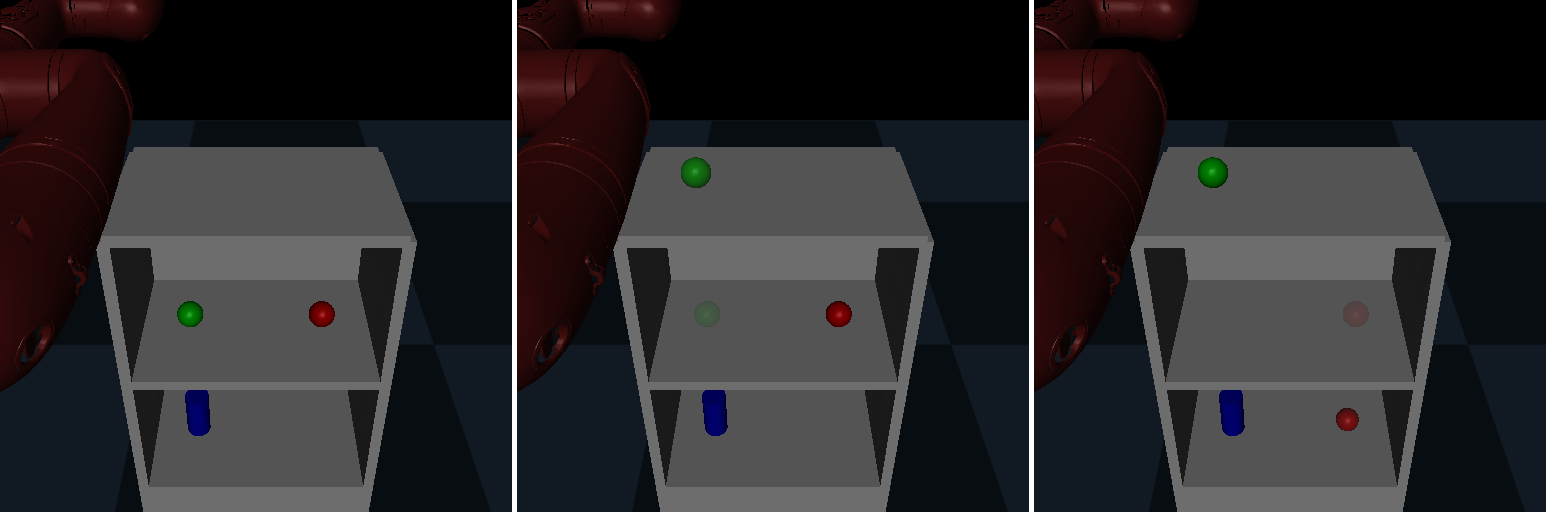} & \includegraphics[scale=\imgscale]{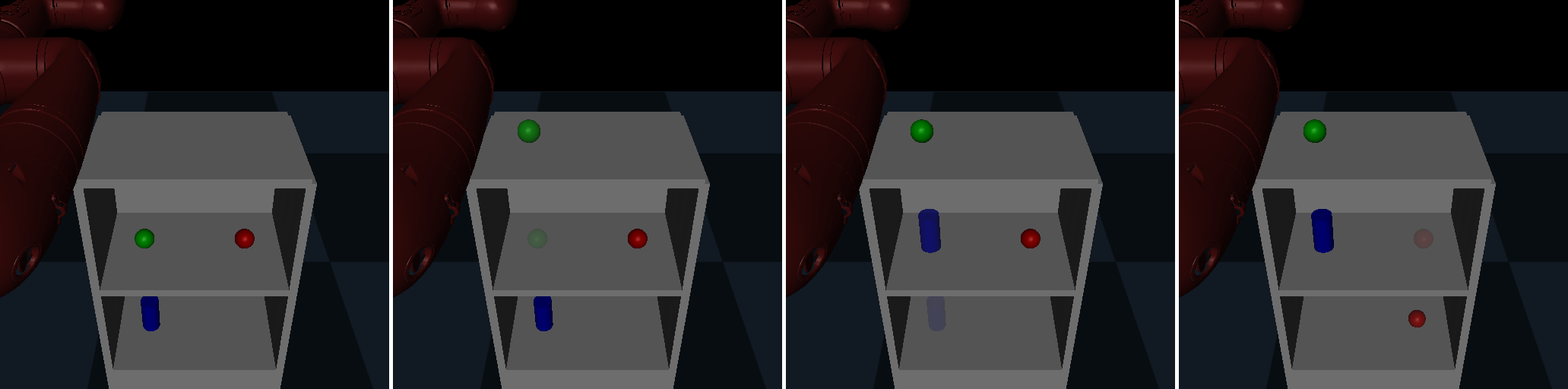} \\ 
                              &	Example Source Task\Bstrut 							& Example Target Task								\\ \hline
6.					          & \textbf{Modify a step}\Tstrut                                   &                                                  \\
                              & \includegraphics[scale=\imgscale]{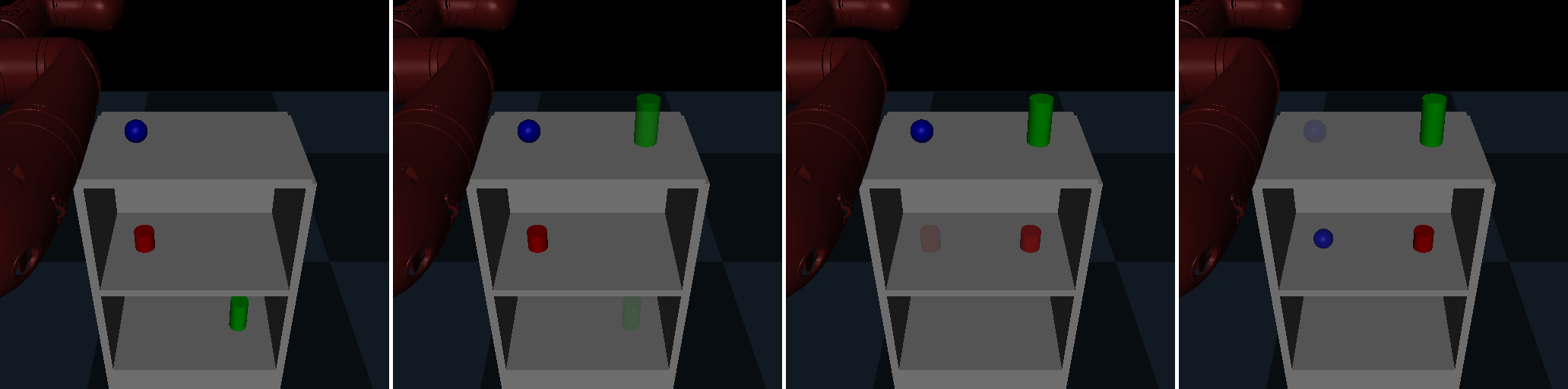} & \includegraphics[scale=\imgscale]{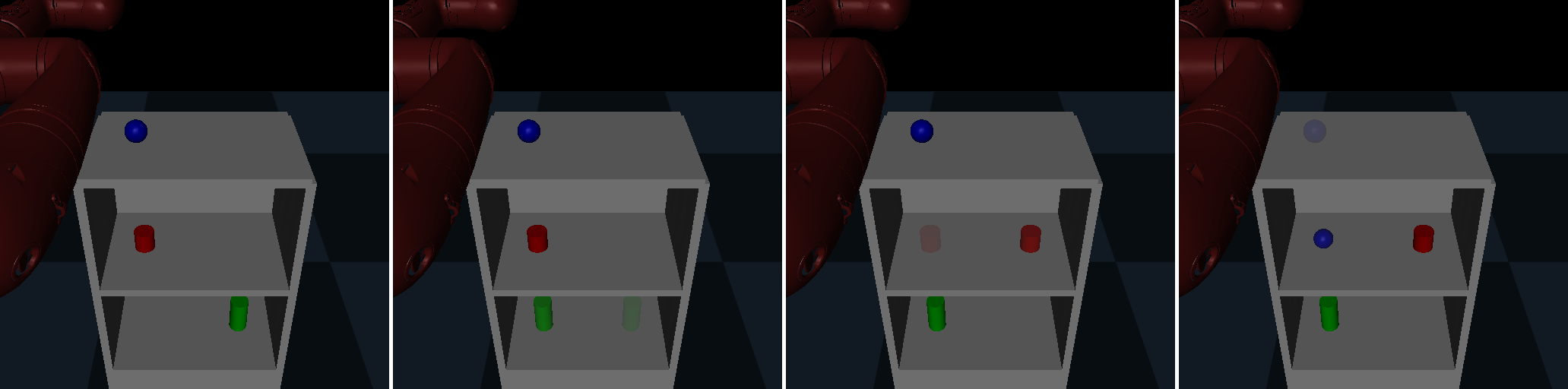} \\ 
                              &	Example Source Task\Bstrut 							& Example Target Task								\\ \hline   
\end{tabular}
\label{tbl:adaptations}
\end{table}

\begin{figure}
    \centering
    \includegraphics[width=\textwidth]{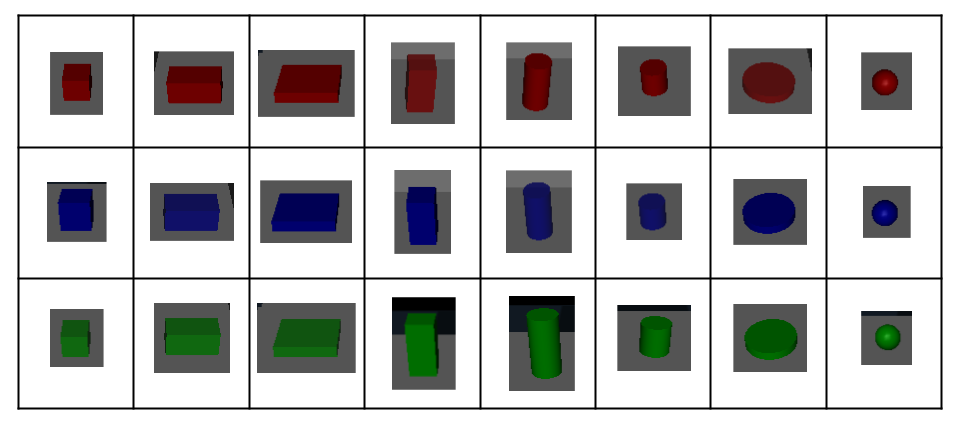}
    \caption{Objects in the Organizer Environment}
    \label{fig:objects}
\end{figure}

Figure~\ref{fig:objects} shows all the objects in the Organizer environment.
Table~\ref{tbl:adaptations} shows an example for each type of adaptation described in Section~\ref{sec:lang}.

\section{Compute Resources}
\label{sec:compute}

The experiments were primarily run on a 64-CPU machine, with a Quadro RTX 6000 GPU. Each experimental run took on average 5 hours to complete.

\section{Additional Experiments}
\label{sec:add-expts}

\begin{table}[]
\caption{Success rates (\%) when using varying amounts of synthetic and natural language data to train LARVA. The row labels show the number of natural language examples used
while the column labels show the number of synthetic language examples used.}
\subfloat[Synthetic language test set]{
\small
\begin{tabular}{rrrrr}
\hline
\multicolumn{1}{c}{\multirow{2}{*}{\textbf{\# natural}}} & \multicolumn{4}{c}{\textbf{\# synthetic}}                   \\ \cline{2-5} 
\multicolumn{1}{c}{}                                     & \textbf{0} & \textbf{7,500} & \textbf{15,000} & \textbf{30,000} \\ \hline
\textbf{0}                                                 &   -         & 83.8          & 93.2           & 97.3           \\ \hline
\textbf{750}                                               & 3.0        & 85.8          & 93.4           & 97.2           \\ \hline
\textbf{1,500}                                              & 5.9        & 85.8          & 91.9           & 96.8           \\ \hline
\textbf{3,000}                                              & 30.1       & 88.1          & 93.6           & 97.7           \\ \hline
\end{tabular}
}
\hfill
\subfloat[Natural language test set]{
\small
\begin{tabular}{rrrrr}
\hline
\multicolumn{1}{c}{\multirow{2}{*}{\textbf{\# natural}}} & \multicolumn{4}{c}{\textbf{\# synthetic}}                   \\ \cline{2-5} 
\multicolumn{1}{c}{}                                     & \textbf{0} & \textbf{7,500} & \textbf{15,000} & \textbf{30,000} \\ \hline
\textbf{0}         & -      & 48.7 & 46.3  & 51.0  \\ \hline
\textbf{750}       & 1.7  & 60.7 & 58.7  & 65.7  \\ \hline
\textbf{1,500}      & 4.3  & 63.0 & 64.0  & 73.3  \\ \hline
\textbf{3,000}      & 29.3 & 68.7 & 72.0  & 73.3  \\ \hline
\end{tabular}
}
\label{tbl:learning-curves}
\end{table}
In order to better understand the amount of data needed, we trained LARVA with varying amounts of synthetic and natural language training examples (using value prediction).
The results are summarized in Table~\ref{tbl:learning-curves}.

Unsurprisingly, on the synthetic language test set, the number of natural language examples only makes a difference when the number of synthetic language examples in the training set is small. 
The results on the natural language test set are more informative. 
In particular, if no natural language examples are used for training, the model is only able to successfully complete about 50\% of the tasks, even as the amount of synthetic language data is increased. 
Furthermore, using 1,500 natural language examples instead of 3,000 with 30,000 synthetic language examples results in a comparable performance as the full set.
Similarly, halving the amount of synthetic language data (i.e. 15,000 examples instead of 30,000) when using the full natural language set results in only a small reduction in performance. However, it is clear that some amount of natural training data is needed to successfully generalize to natural language test cases.

These results suggest that using additional synthetic language or natural language data compared to our full set will likely not result in a significant performance improvement, and thus, improving the performance when using natural language requires filtering out low quality natural language data, and using richer models.

\section{Societal Impacts}
\label{sec:broader-impacts}

Our approach is a step towards building agents that can be trained more conveniently by non-experts, which is a crucial property for AI that can work alongside humans to accomplish everyday tasks.

While such AI can have wide-reaching positive impact, for instance, in taking care of the elderly or differently-abled, and accomplishing simple tasks in small-scale businesses freeing up valuable human time,
such technology can also be used for unethical objectives.
First, such a system could be taught malicious behavior more conveniently, using demonstrations and language. Thus, the downstream system that executes these behaviors in the real-world (e.g. a mobile robot) must be designed with safety as a top priority. Further, there should be a system in place (either manual or automated) which should filter out potentially malicious data, both while training and predicting from a model like LARVA. One way to do this could be to check if the language contains any words that could cause harm (e.g. ``attack'').
Second, research in natural language processing is largely based on English. Care must be taken to ensure that as these systems are scaled, support for other languages is provided, as well as variations in accents, social backgrounds, etc. are handled, so that the technology can be used to benefit all sections of the society.

\end{document}